\pdfoutput=1
\documentclass[11pt]{article}

\usepackage{emnlp2023}

\usepackage{times}
\usepackage{latexsym}

\usepackage[T1]{fontenc}
\usepackage[utf8]{inputenc}

\usepackage{microtype}

\usepackage{refstyle}
\usepackage{inconsolata}
\usepackage{tcolorbox}
\usepackage{amsmath,amsfonts,amssymb}
\usepackage{float}
\usepackage{booktabs}
\usepackage{graphics}
\usepackage{multicol}
\usepackage{multirow}
\usepackage{colortbl}
\definecolor{GREEN}{RGB}{84,130,53}
\newcommand{\colorit}{\cellcolor{green!15}}
\newcommand{\coloritt}{\cellcolor{orange!15}}

\usepackage{pifont}

\definecolor{ballblue}{rgb}{0.13, 0.67, 0.8}
\definecolor{azure1}{rgb}{0.0, 0.5, 1.0}
\definecolor{uclablue}{rgb}{0.33, 0.41, 0.58}
\definecolor{ultramarine}{rgb}{0.07, 0.04, 0.56}
\definecolor{yaleblue}{rgb}{0.06, 0.3, 0.57}

\usepackage{cleveref}
\crefformat{section}{\S#2#1#3}
\crefformat{subsection}{\S#2#1#3}
\crefformat{subsubsection}{\S#2#1#3}
\crefrangeformat{section}{\S\S#3#1#4 to~#5#2#6}
\crefmultiformat{section}{\S\S#2#1#3}{ and~#2#1#3}{, #2#1#3}{ and~#2#1#3}
\Crefformat{figure}{#2Fig.~#1#3}
\Crefmultiformat{figure}{Figs.~#2#1#3}{ and~#2#1#3}{, #2#1#3}{ and~#2#1#3}
\Crefformat{table}{#2Tab.~#1#3}
\Crefmultiformat{table}{Tabs.~#2#1#3}{ and~#2#1#3}{, #2#1#3}{ and~#2#1#3}
\Crefformat{appendix}{\S#2#1#3}
\crefformat{algorithm}{Alg.~#2#1#3}

\title{Context-faithful Prompting for Large Language Models}

\author{Wenxuan Zhou$^{1}$, Sheng Zhang$^{2}$, Hoifung Poon$^{2}$, Muhao Chen$^{1}$\\
$^{1}$University of Southern California $^{2}$Microsoft Research \\ 
$^{1}${\{zhouwenx,muhaoche\}@usc.edu}\; $^{2}${{\{zhang.sheng,hoifung\}@microsoft.com}}}

\begin{document}
\maketitle
\begin{abstract}
Large language models (LLMs) encode parametric knowledge about world facts and have shown remarkable performance in knowledge-driven NLP tasks.
However, their reliance on parametric knowledge may cause them to overlook contextual cues, leading to incorrect predictions in context-sensitive NLP tasks (e.g., knowledge acquisition tasks).
In this paper, we seek to assess and enhance LLMs' contextual faithfulness in two aspects: knowledge conflict and prediction with abstention.
We demonstrate that LLMs' faithfulness can be significantly improved using carefully designed prompting strategies.
In particular, we identify opinion-based prompts and counterfactual demonstrations as the most effective methods.
Opinion-based prompts reframe the context as a narrator's statement and inquire about the narrator's opinions, while counterfactual demonstrations use instances containing false facts to improve faithfulness in knowledge conflict situations.
Neither technique requires additional training.
We conduct experiments on three datasets of two standard NLP tasks, machine reading comprehension and relation extraction, and the results demonstrate significant improvement in faithfulness to contexts.\footnote{Code and data are released at \url{https://github.com/wzhouad/context-faithful-llm}.}
\end{abstract}

\section{Introduction}
Large language models (LLMs; \citealt{brown2020language,weifinetuned,chowdhery2022palm,chung2022scaling}) have made remarkable advances in solving various NLP problems, particularly in (context-free) knowledge-driven tasks such as question answering~\cite{joshi-etal-2017-triviaqa,kwiatkowski-etal-2019-natural} and commonsense reasoning~\cite{clark2018think,mihaylov-etal-2018-suit}.
Without external context, LLMs can answer factual questions and achieve comparable results to supervised approaches~\cite{brown2020language,weifinetuned}, indicating that LLMs encode \emph{parametric knowledge} about open-world facts.

\begin{figure}
    \centering
    \includegraphics[width=0.9\linewidth]{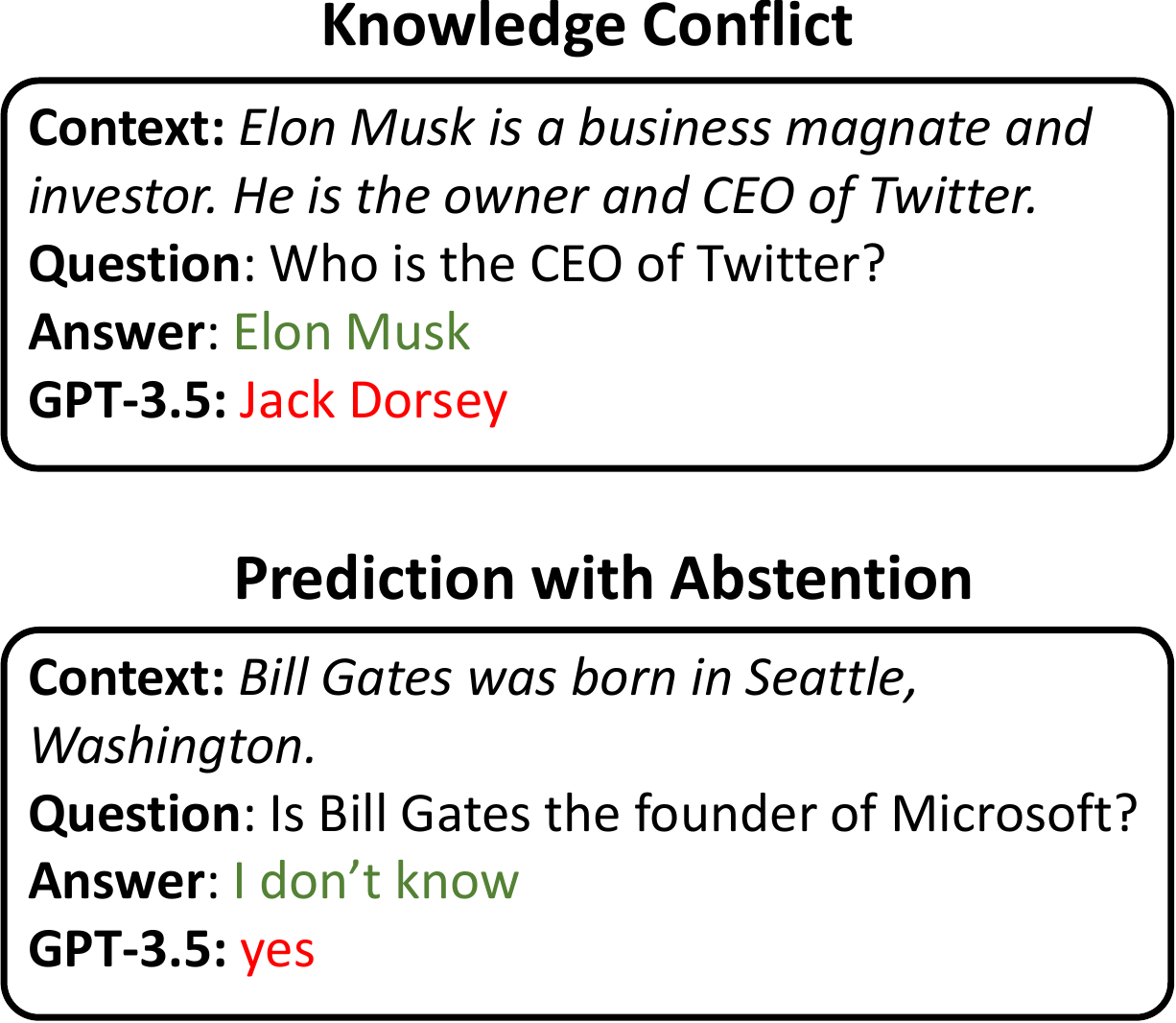}
    \caption{Examples of knowledge conflict and prediction with abstention. LLMs may ignore the provided context and make unfaithful predictions based on their parametric knowledge before Q4 2021.}
    \label{fig:intro}
    %\vspace{-1em}
\end{figure}

Although parametric knowledge can be beneficial for knowledge-driven tasks, overly relying on it can cause problems in context-specific NLP tasks.
First, LLMs may encode misconceptions~\cite{lin-etal-2022-truthfulqa} or obsolete facts~\cite{lazaridou2021mind,liska2022streamingqa,kasai2022realtime}, in which case we expect LLMs to update their predictions when provided with relevant context.
Second, when using LLMs for knowledge acquisition tasks such as machine reading comprehension (MRC;~\citealt{clark-etal-2019-boolq,rajpurkar-etal-2016-squad}) and information extraction (IE;~\citealt{sang2003introduction,zhang-etal-2017-position,zhou-chen-2022-improved,lu-etal-2022-summarization}), LLMs should always extract the \emph{knowledge in context} instead of relying solely on their parametric knowledge.
In such context-specific application scenarios, we expect LLMs to make decisions faithful to the context and avoid simply parroting answers from pretraining.
However, studies have discovered that LLMs can overlook or ignore context~\cite{kasai2022realtime,li2022large,si2022prompting}, posing a significant challenge for their application in these scenarios.

In this paper, we aim to investigate techniques for improving the faithfulness of LLMs in context-specific NLP tasks.
Conceptually, faithfulness is not simply about how much accuracy the model can offer.
Instead, it should concern the validity and reliability of its extraction process. 
Specifically, when there is decision-related information (e.g., a concept or a relation) to extract, a faithful LLM should \emph{genuinely} induce what is described in the context but not give \emph{trivial guesses} based on parametric knowledge or statistical biases.
Besides, when no known decision-related information is described in the context, the model should \emph{selectively} abstain from predicting.
%Since faithfulness is a broad topic, we narrow our focus to two specific facets
Accordingly, to provide a realistic assessment of LLMs in terms of faithfulness, we narrow our focus to two sub-problems, namely entity-based knowledge conflict~\cite{longpre-etal-2021-entity,wang-etal-2022-rely} and prediction with abstention~\cite{rajpurkar-etal-2018-know}, examples of which are shown in~\Cref{fig:intro}.
In cases of knowledge conflict, where the given context contains facts different from the pretraining data, LLMs need to return the facts locally described in the context instead of the globally memorized ones.
For example, in~\Cref{fig:intro}, text-davinci-003 identifies \textit{Jack Dorsey} instead of \textit{Elon Musk} as the CEO of Twitter, based on its pretrained data before Q4 2021.
In cases of prediction with abstention, where the context does not provide information to answer the questions, LLMs should abstain from making predictions and notify the users, rather than answering the questions that become a trivial guess.
For example, in~\Cref{fig:intro}, when asked about the founder of Microsoft based on an irrelevant context, LLMs should admit that, from here, they cannot infer the answer.

We present various prompting strategies to improve the faithfulness of LLMs, including designing effective prompts and choosing appropriate in-context demonstrations.
We find that constraining the scope of questions to the context by adding phrases (e.g., based on the given context) or natural language instructions improve faithfulness in both facets.
Particularly, we find that reformulating the context and questions to opinion-based question-answering problems~\cite{guptaamazonqa19,bjerva-etal-2020-subjqa}, where the context is expressed in terms of a narrator's statement, and the question asks about this narrator's opinion, delivers the most gains.
Additionally, we find that adding counterfactual demonstrations to prompts improves faithfulness in the aspect of knowledge conflict, while using the original (factual) demonstrations leads to limited or negative effects.
Finally, combining both techniques delivers the largest gain than using each one independently.

We evaluate our methods based on three datasets, including Re-TACRED~\cite{stoica2021re} for relation extraction, and natural questions~\cite{kwiatkowski-etal-2019-natural} and RealTime QA~\cite{kasai2022realtime} for MRC.
We find that the proposed strategies can largely improve faithfulness, e.g., reducing the memorization ratio\footnote{The percentage of times that LLMs return memorized answers versus answers in the context \cite{longpre-etal-2021-entity}.} of text-davinci-003 from 35.2\% to 3.0\% on natural questions.
Additionally, we evaluate our methods across LLMs of different scales, finding that larger LLMs are more likely to update memorized answers than smaller ones, both with and without the application of our methods.
% \muhao{become more faithful than smaller models after applying our approach?}.

% \section{Problem Setting}
% In order to evaluate LLMs' faithfulness on context, we need to disentangle the knowledge from the context and the parametric knowledge learned in pretraining.
% We adopt the knowledge conflict setting~\cite{longpre-etal-2021-entity} in MRC.
% Specifically, given an instance $(q, a, c)$ with query $q$, answer $a$, and context $c$, the knowledge conflict framework produces a new instance $(q, a', c')$, where $a$ and all of its occurrences in $c$ are replaced with a new answer $a'$, producing new context $c'$.
% In this case, a faithful LLM needs to update its predicted answer to $a'$ instead of returning the original answer $a$.

\section{Related Work}

We discuss two topics of related work that are closely relevant to this work.

\smallskip
\noindent\textbf{Knowledge conflicts.} LLMs~\cite{brown2020language,weifinetuned,chowdhery2022palm} have shown promising results in closed-book QA tasks, indicating their ability to memorize facts about the world.
However, as the world is constantly evolving, memorized facts may become outdated~\cite{lazaridou2021mind,liska2022streamingqa,kasai2022realtime}, emphasizing the need to update LLMs' predictions with new facts.
To address this challenge, some studies~\cite{zhu2020modifying,de-cao-etal-2021-editing,mitchellfast,meng2022locating,meng2022mass} have explored ways to identify and edit the facts stored in model parameters.
% For example, \citet{meng2022mass} identify the MLPs that have a causal effect on factual knowledge recall in LLMs and edit the parameters in these critical MLPs to update the knowledge.
% Their method can update up to thousands of memorized facts with little impairment to others.
However, it remains unclear whether memory editing methods allow sufficient capacity to encompass all new factual knowledge.
Another promising direction is to augment LLM prompting with external context containing relevant knowledge~\cite{lazaridou2022internet,izacard2022few,khattab2022demonstrate}.
Coupled with retrieval systems~\cite{karpukhin-etal-2020-dense,santhanam-etal-2022-colbertv2,gao-callan-2022-unsupervised}, such methods have the potential to update LLMs with large amounts of new facts.
However, such methods face the challenge that LLMs may persist with the memorized facts and ignore the provided context~\cite{longpre-etal-2021-entity}.
To tackle this challenge, recent works~\cite{neeman2022disentqa,li2022large} finetune LLMs on counterfactual contexts, where the original facts are replaced with counterfactual ones.
They find that such finetuning processes can effectively improve the LLMs' utilization of contexts.
In this study, we propose a novel approach using prompting to improve context faithfulness in LLMs without additional finetuning, which offers a more general and cost-effective method for LLMs.

\smallskip
\noindent\textbf{Prediction with abstention.}
Selective prediction with abstention~\cite{chow1970optimum,fumera2002support,cortes2016learning} is an important problem in trustworthy AI.
When models are uncertain about their predictions, it is critical that they should admit the uncertainty instead of returning incorrect predictions.
Selective prediction may be adopted in different scenarios, such as when instances are close to the decision boundary~\cite{gal2016dropout,lakshminarayanan2017simple,xin-etal-2021-art}, or when instances are from different domains to training~\cite{hendrycksbaseline,hendrycks-etal-2020-pretrained,zhou-etal-2021-contrastive}.
In the scope of context-specific NLP, abstention is preferred when the context is irrelevant to the question.
For example, SQuAD 2.0~\cite{rajpurkar-etal-2018-know} introduces unanswerable questions to extractive MRC, while~\citet{yatskar-2019-qualitative} finds it focused on questions of extreme confusion and thus is less relevant to the focus of our study.
CoQA~\cite{reddy-etal-2019-coqa} and QuAC~\cite{choi-etal-2018-quac} introduce unanswerable questions to conversational question answering.
RealTime QA~\cite{kasai2022realtime} finds that GPT-3 still generates outdated answers when provided with irrelevant documents.
To address the problem, \citet{neeman2022disentqa} propose the answerability augmentation where LLMs should predict \textit{Unanswerable} when presented with an empty or randomly sampled document.
Several other work employ variants of confidence calibration techniques to encourage the NLP model to avoid giving a high confidence on any decisions when encountering a case to abstain \cite{wang-etal-2023-extracting,wang-etal-2022-rely}, which however request white-box accessibility of the incorporated models.
We tackle this problem with a part of our prompting method, which we find to significantly enhance the LLMs' ability to make selective predictions without need re-calibration or white-box accessibility of the model.

\section{Method}
We focus on context-specific NLP tasks.
The input of these tasks is formulated as $(c, q)$ for free-form generation tasks, where $c$ is the context and $q$ is the question, or $(c, q, o)$ for %multi-choice tasks
tasks with close decision spaces (e.g., multi-choice tasks), where $o$ is the set of decisions/choices.
The desired output can be either a free-form text or a choice.
% \muhao{@wenxuan: overview: specify what techniques are you proposing? }
We solve these tasks by prompting LLMs and %seek to improve the faithfulness of LLMs in two ways, namely prompt engineering and demonstrations.
study ways of designing prompting templates and demonstrations that are dedicated to improving the faithfulness of LLMs.
Specifically, we find two proposed methods, opinion-based prompts and counterfactual demonstrations, to be the most effective ones.
Our methods only change the prompts without finetuning the LLMs~\cite{longpre-etal-2021-entity,li2022large,neeman2022disentqa}, targeting a more general and affordable solution.
% \muhao{@zhouwenx: btw the Method section hasn't mentioned anythin related to prediction with abstention.}

\subsection{Opinion-based Prompting}
\label{ssec:prompting_templates}
% \muhao{better not call it prompt engineering. just call it opinionating or some more technically specific name; prompt engineering looks pretty trivial.}

% \muhao{need to describe what are the key techniques at the beginning (e.g. instruction-based constraints and opinionated context.) otherwise it's hard for reviewers to catch the key ideas.}
Given an input $(c, q, o)$, we begin with the following \emph{base} prompting template:\footnote{Options only apply to multiple-choice tasks and are removed in free-form text generation tasks.}
\begin{tcolorbox}[title=Base prompt]
\small
$\{c\}$ Q: $\{q\}$? Options: $\{o\}$ A:
\end{tcolorbox}
Here, $\{.\}$ serves as a placeholder to be filled with specific content during prompting.
We investigate two types of prompting templates for context-specific NLP, namely \emph{opinion-based} prompts and \emph{instructed} prompts.
Opinion-based prompts transform original questions into opinion-seeking questions, which naturally demand more attention to the context.
Instructed prompts, on the other hand, explicitly instruct LLMs to read the context by natural language.
Details of these templates are discussed in the remaining section.

\smallskip
\noindent \textbf{Opinion-based prompts.}
We propose to transform the context to a narrator's statement and the question to enquire about the narrator's opinion in this statement.
This approach is motivated by our own cognitive process for answering different types of questions.
When answering questions that seek factual information, we can often rely on our own memory and answer without needing to refer to the context, as these questions typically have only one correct answer.
However, when questions are seeking opinions from someone else (in this context, the narrator), it is important to comprehend the narrator's words before answering the questions, as opinions may vary from person to person.
Besides, as opinions are inherently subjective and can be influenced by many factors such as personal experiences and beliefs, opinion-seeking questions are sometimes difficult to answer solely based on the narrator's statement compared to a fact-seeking question that typically has definite and verifiable answer(s).
As a result, transforming factual questions into opinion-seeking questions can lead to more attention to the context, as memorized answers alone may not suffice.
It also helps the model more selectively predict under cases where contexts do not describe answers. 
Both factors lead to improved faithfulness with LLMs.
The \emph{opinion}-based prompting template is as follows:
\begin{tcolorbox}[title=Opinion-based prompt]
\small
Bob said, ``$\{c\}$'' Q: $\{q\}$ in Bob's opinion? Options: $\{o\}$ A:
\end{tcolorbox}
Throughout our experiments, we consistently use Bob to represent the narrator for the context, although other names could be utilized as well.

\smallskip
\noindent \textbf{Instructed prompts.}
We also explicitly instruct LLMs to read context by natural language.
We start by extending questions in prompts with attributive phrases such as ``based on the given text'', leading to the following \emph{attributed} prompting template:
\begin{tcolorbox}[title=Attributed prompt]
\small
$\{c\}$ Q: $\{q\}$ based on the given text? Options:\{o\} A:
\end{tcolorbox}

We also augment the prompts with natural language instructions.
Since manually writing instructions can be laborious and often fails to account for the compatibility between instructions and LLMs, we leverage automatic prompt engineering~(APE; \citealt{zhou2022large}) to generate the prompts.
Using a few instances and their desired outputs as demonstrations, APE uses LLMs to automatically generate candidate instructions and select the best one based on the results on a dev set~(see Appx.~\Cref{ssec:ape_instructions} for generated instructions).
% \muhao{may need to provide an example for APE-generated instruction.}
We then use the following \emph{instruction}-based prompting template:
\begin{tcolorbox}[title=Instruction-based prompt]
\small
Instruction: \{\textit{Instruction}\} $\; \{c\}$ Q: $\{q\}$? Options: $\{o\}$ A:
\end{tcolorbox}

Experiments show that all prompting templates perform better than the base prompting template.
Specifically, opinion-based prompts outperform instructed prompts in both knowledge conflict and prediction with abstention facets, and combining these two prompting methods results in the most significant improvements.

\subsection{Counterfactual Demonstration}\label{ssec:demonstration}
Using demonstrations is a standard way to perform few-shot inference on LLMs~\cite{brown2020language}.
To enhance the faithfulness of language models in knowledge conflict scenarios, previous studies~\cite{li2022large,neeman2022disentqa} propose to finetune the models using counterfactual instances, where the facts in the context are substituted with false ones, and the model learns to update its predictions accordingly.
Following this strategy, we propose to use counterfactual instances as demonstrations for LLMs.
To do so, we start with a labeled set of counterfactual instances and a test instance and then use KATE~\cite{liu-etal-2022-makes} to retrieve the most relevant counterfactual instances as demonstrations.
We encode both the test instance and counterfactual instances with RoBERTa$_\text{nli+sts-b}$~\cite{liu2019roberta,reimers-gurevych-2019-sentence} and select the top counterfactual instances based on cosine similarity.
As a part of our analysis, we also experimented with using the original (factual) instances as demonstrations but found this approach to underperform counterfactual demonstrations and sometimes even zero-shot inference.

\begin{table*}[!t]
    \centering
    %\scalebox{0.82}
    {
    \small
    \begin{tabular}{llcccccccc}
    \toprule
     &\multirow{2}{*}{GPT-3.5}& \multicolumn{4}{c}{MRC}& \multicolumn{4}{c}{RE} \\
     && $p_s\uparrow$& $p_o\downarrow$& $M_R\downarrow$& EM$\uparrow$&$p_s\uparrow$& $p_o\downarrow$& $M_R\downarrow$& $F_1$$\uparrow$\\
     \midrule
     \multirow{5}{*}{\rotatebox[origin=c]{90}{Zero-shot}}&Base& 59.0& 32.1& 35.2& 6.2& 73.9& 21.5& 22.5& 81.0\\
     &Attr&71.9& 14.4&16.6& \coloritt 29.6&72.4& 23.6&24.6& 80.0 \\
     &Instr& 74.2& 16.0& 17.7& 27.1& 75.8& \coloritt 15.6& \coloritt 17.1& 81.6 \\
     &Opin& \colorit 79.4&\coloritt 9.8& \coloritt 11.0& 24.9& \coloritt 76.0&19.6& 20.5& \coloritt 82.9 \\
     &Opin + Instr& \coloritt 79.1& \colorit 7.9& \colorit 9.1& \colorit 48.6& \colorit79.4& \colorit 15.0& \colorit15.9& \colorit84.7 \\
     \midrule
     \multirow{5}{*}{\rotatebox[origin=c]{90}{Original}}&Base& 43.3& 49.4& 53.3& 35.1& 76.2& 19.8& 20.6& 83.3\\
     &Attr&54.1& 37.7&41.0& 45.5&76.5& 19.7& 20.5& 83.7 \\
     &Instr& 54.6&  37.7& 40.8& 45.8& \coloritt 77.3& \coloritt 18.4& \coloritt 19.2& \coloritt 84.2\\
     &Opin& \coloritt 60.6& \coloritt 28.7& \coloritt 32.1& \coloritt 51.1& 76.8&\coloritt 18.4& 19.3& 83.8 \\
     &Opin + Instr& \colorit64.7& \colorit26.8& \colorit29.3& \colorit53.8& \colorit78.2& \colorit17.1& \colorit17.9&\colorit 84.9 \\
     \midrule
     \multirow{5}{*}{\rotatebox[origin=c]{90}{Counter}}&Base& 86.9& 6.5& 7.0& 80.2& 78.7& 13.7& 14.8& 83.9\\
     &Attr&89.1& 4.6&4.9& 83.0& \coloritt 79.7& 13.0& 14.0& 84.3 \\
     &Instr& 86.2& 6.3& 6.8& 80.1& 78.0& \coloritt 12.8& 14.1& 82.9\\
     &Opin& \coloritt 90.1&\coloritt 3.7& \coloritt 3.9& \coloritt 84.3& \coloritt 79.7& \coloritt 12.8& \coloritt 13.8& \coloritt 84.4 \\
     &Opin + Instr& \colorit\textbf{90.9}& \colorit\textbf{2.8}& \colorit\textbf{3.0}& \colorit\textbf{85.2}& \colorit\textbf{80.0}& \colorit\textbf{10.5}&\colorit\textbf{11.6}&\colorit\textbf{85.1} \\
     \bottomrule
    \end{tabular}
    }
    \caption{Results (in \%) on GPT-3.5-175B in the knowledge conflict setting. The overall best results are highlighted in \textbf{bold}. The best and the second best results in each setting are highlighted in \colorbox{green!15}{green} and \colorbox{orange!15}{orange}, respectively.}
    \label{tab:main_result}
    %\vspace{-1em}
\end{table*}

\begin{table*}[!t]
    \centering
    %\scalebox{0.82}
    {
    \small
    \begin{tabular}{llcccccccc}
    \toprule
     &\multirow{2}{*}{LLama-2}& \multicolumn{4}{c}{MRC}& \multicolumn{4}{c}{RE} \\
     && $p_s\uparrow$& $p_o\downarrow$& $M_R\downarrow$& EM$\uparrow$&$p_s\uparrow$& $p_o\downarrow$& $M_R\downarrow$& $F_1$$\uparrow$\\
     \midrule
     \multirow{5}{*}{\rotatebox[origin=c]{90}{Zero-shot}}&Base& 50.8& 40.9& 44.6& 3.5& 15.3& 67.6& 81.6& 12.8\\
     &Attr&66.2& 23.8&26.4& 4.7& 13.2& 66.5& 83.5& 10.9\\
     &Instr& \coloritt 77.7& 19.7& 20.2& \colorit 27.0& 19.6& \coloritt 9.2& \coloritt 75.1& \coloritt 13.2\\
     &Opin& 74.6& \coloritt 14.6& \coloritt 16.4& 9.4& \coloritt 20.7& 63.4& 75.4& 14.4\\
     &Opin + Instr& \colorit 77.8& \colorit 13.9& \colorit 15.1& \coloritt 13.7& \colorit 21.6& \colorit 57.9& \colorit 72.8& \colorit 11.8\\
     \midrule
     \multirow{5}{*}{\rotatebox[origin=c]{90}{Original}}&Base& 56.7& 39.7& 41.1& 19.4 & 27.6& 62.3& 69.4& 9.4 \\
     &Attr&61.7& 34.5&35.9& 25.2& 29.4& 58.9& 66.7& 11.2\\
     &Instr& 59.4& 35.7& 37.5& \coloritt 25.5& \coloritt 34.6& \coloritt 53.6& \coloritt 60.8& \colorit 13.2 \\
     &Opin& \coloritt 67.1& \coloritt 32.1& \coloritt 32.4& 18.5& 32.2& 57.1& 63.9& 10.9\\
     &Opin + Instr& \colorit 70.6& \colorit 26.8& \colorit 27.5& \colorit 27.6& \colorit 35.7& \colorit 51.3& \colorit 59.0& \coloritt 11.5\\
     \midrule
     \multirow{5}{*}{\rotatebox[origin=c]{90}{Counter}}&Base& 84.4& 7.8& 8.4& 39.2& 76.3& 14.8& 16.2& 38.9\\
     &Attr&85.9& 7.0&7.6& 44.1& \coloritt 76.5& \coloritt 14.2& 15.7& \coloritt 39.5\\
     &Instr& 85.5& 6.7& 7.3& \coloritt 47.1& 76.0& 14.4& 15.9& 37.3\\
     &Opin& \coloritt 86.7&\coloritt 6.2& \coloritt 6.7& 38.1& 76.3& \colorit \textbf{13.8}& \colorit \textbf{15.4}& \colorit \textbf{41.7}\\
     &Opin + Instr& \colorit \textbf{88.1}& \colorit \textbf{4.9}& \colorit \textbf{5.2}& \colorit \textbf{49.6}& \colorit \textbf{77.3}& \coloritt 14.2& \coloritt 15.5& 36.9\\
     \bottomrule
    \end{tabular}}
    \caption{Results (in \%) on LLama-2-7B-chat in the knowledge conflict setting. The overall best results are highlighted in \textbf{bold}. The best and the second best results in each setting are highlighted in \colorbox{green!15}{green} and \colorbox{orange!15}{orange}, respectively.}
    \label{tab:main_result_llama2}
    %\vspace{-1em}
\end{table*}

\section{Experiments}
This section presents our experimental setups~(\Cref{ssec:experimental_settings}) for the evaluation of the proposed methods concerning two aspects of faithfulness: knowledge conflict~(\Cref{ssec:knowledge_conflict}) and prediction with abstention~(\Cref{ssec:abstention}).
We provide additional analysis~(\Cref{ssec:analysis}) on results across different model sizes and results on the original datasets.
We also show examples of prompts and LLMs' outputs in the case study~(\Cref{ssec:case_study}).

\subsection{Experimental Setup}
\label{ssec:experimental_settings}
Our experiments are conducted using the InstructGPT model (text-davinci-003, 175B parameters) and LLama-2-7B-chat~\cite{touvron2023llama}.
We use the base prompt as our baseline, and compare it against the proposed prompting templates described in~\Cref{ssec:prompting_templates}, including attributed prompt (\textsc{Attr}), instruction-based prompt (\textsc{Instr}), opinion-based prompt (\textsc{Opin}), and the combination of opinion-based prompt and instruction-based prompt (\textsc{Opin + Instr}).
We evaluate the effectiveness of these templates in both zero-shot and few-shot settings (with demonstrations).
\subsection{Knowledge Conflict}
\label{ssec:knowledge_conflict}

\smallskip
\noindent
\textbf{Datasets.} We evaluate in the knowledge conflict setting using counterfactual datasets that contain incorrect facts, which can %cause knowledge conflicts with LLMs.
conflict with what the LLM has memorized.
We use two datasets based on real-world texts: natural questions~\cite{kwiatkowski-etal-2019-natural} for MRC and Re-TACRED~\cite{stoica2021re} for relation extraction~(RE).
To create counterfactuals, we adopt the framework proposed by~\citet{longpre-etal-2021-entity}, which modifies the context to support a counterfactual answer.
Specifically, for MRC, we follow~\citet{longpre-etal-2021-entity} and replace the gold entity answer in the context with a randomly sampled entity of the same entity type from the corpus.
For RE, we first randomly sample a context that has the entity mentions of the same type but different relations from the original one, and then insert the original entities into the sampled context.
In this scenario, a faithful LLM should update its prediction to the new answer instead of returning the original one.
Moreover, to measure LLMs' ability to update answers, we need to ensure that they have memorized the knowledge of the original answers in the first place.
Therefore, we only evaluate LLMs on a subset of instances on which these models can correctly predict the original answers without additional contexts.
% This process left us with 2,773 test instances on natural questions and 4,353 test instances on Re-TACRED.

\begin{table*}[!t]
    \centering
    %\scalebox{0.85}
    {
    \small
    \begin{tabular}{llcccccccc}
    \toprule
     &\multirow{2}{*}{Method}&\multicolumn{3}{c}{GPT-3.5}& \multicolumn{4}{c}{LLama-2} \\
     &&\multicolumn{2}{c}{Acc$\uparrow$}& \multirow{2}{*}{Brier$\downarrow$}& \multicolumn{3}{c}{Acc$\uparrow$}& \multirow{2}{*}{Brier$\downarrow$}\\
     && NoAns& All&& HasAns& NoAns& All& \\
     \midrule
     \multirow{5}{*}{\rotatebox[origin=c]{90}{Zero-shot}}&Base& 30.6 & 68.5& 29.4& \coloritt 88.7& 14.3& 55.9& 30.0\\
     &Attr& 65.3&84.4& 14.6& 87.1& 16.3& 55.9& 29.7\\
     &Instr&  81.6& 91.7& 7.7& \colorit \textbf{91.9}& 26.5& \coloritt 63.1& \coloritt 27.4\\
     &Opin&  \coloritt 83.3&  \coloritt 92.6& \coloritt 6.6& 85.5& \coloritt 30.6& 61.3& \colorit 27.8\\
     &Opin + Instr&  \colorit 87.8& \colorit94.4& \colorit5.2& \coloritt 88.7& \colorit 36.7& \colorit 65.8& \coloritt 27.4 \\
     \midrule
     \multirow{5}{*}{\rotatebox[origin=c]{90}{Few-shot}}&Base&73.5&88.2& 11.2& \coloritt 56.5& 69.4& 62.2& 27.9\\
     &Attr&81.6&91.9& 8.0& \colorit 59.7& 67.3& 63.1& 27.0\\
     &Instr&85.7& 93.7& 6.1& 51.6& 81.6& \coloritt 64.9& \coloritt 26.6 \\
     &Opin&\coloritt 87.8& \coloritt 94.6& \coloritt 4.1& 50.0& \coloritt 87.8& \colorit \textbf{66.6}& \coloritt 26.6\\
     &Opin + Instr&\colorit \textbf{89.8}& \colorit \textbf{95.5}& \colorit \textbf{3.4}& 43.5& \colorit \textbf{91.8}& \coloritt 64.9& \colorit \textbf{26.0} \\
     \bottomrule
    \end{tabular}}
    \caption{Results (in \%) for GPT-3.5 and LLama-2 on RealTime QA. The overall best results are highlighted in \textbf{bold}. The best and the second best results in each setting are highlighted in \colorbox{green!15}{green} and \colorbox{orange!15}{orange}, respectively. As all prompts achieve perfect accuracy (100\%) on the \textbf{HasAns} subset for GPT-3.5, it is not included in the table.}
    \label{tab:realtime_qa_result}
    %\vspace{-1em}
\end{table*}

\smallskip
\noindent
\textbf{Task setup.}
We use the same set of evaluation metrics as~\citet{longpre-etal-2021-entity}.
Specifically, we measure the frequency that the LLMs' predictions \emph{contain} an exact match of the original answers ($p_o$) and the substituted answers ($p_s$), after both predictions and answers have been normalized by removing stop words and punctuation
To assess the model's reluctance to update its prediction, we use the memorization ratio ($M_R$), which is calculated as $M_R=\frac{p_o}{p_o + p_s}$.
A completely faithful LLM should have an $M_R$ of 0.
We also report task-specific metrics, including exact match~(EM) for MRC and $F_1$ for RE.
For EM, we also use normalized predictions and answers, but the requirement is that the prediction and answer must be exactly the same, rather than just containing the answer.
We conduct experiments in three different settings: zero-shot, demonstration using original instances, and demonstration using counterfactual instances.
We retrieve demonstrations from the original/counterfactual training set, and evaluate LLMs on the counterfactual test set.
In the few-shot setting, we utilize a maximum of 16 demonstration instances, up to the limit of the LLM's context window.

\smallskip
\noindent
\textbf{Results and discussion.}
The results in \Cref{tab:main_result} and \Cref{tab:main_result_llama2} demonstrate that the combination of \textsc{Opin + Instr} prompting and counterfactual demonstrations is generally the most effective.
Compared to the zero-shot base prompts, there is a reduction of 32.2\% in $M_R$ for MRC and a 10.9\% reduction for RE on GPT-3.5.
Similarly, on LLaMA-2-7B-chat, there is a 39.4\% reduction in $M_R$ for MRC and a 57.3\% reduction for RE.
We also find that opinion-based prompts generally perform better than other templates, achieving the second-best results on 17 out of 24 metrics on GPT-3.5, and 9 out of 24 metrics on LLama-2, indicating that LLMs are more faithful to the context when answering opinion-seeking questions.
Combining opinion-based prompts and instruction-based prompts further improves faithfulness, with the best results obtained in 23 out of 24 metrics on GPT-3.5, and 19 out of 24 metrics on LLama-2.

When it comes to few-shot settings, counterfactual demonstrations %generally perform better than other settings.
lead to further improved performance.
Using the original (factual) instances as demonstrations, on the other hand, leads to limited effects or may even impair faithfulness in MRC.
This finding suggests that demonstrations do not always improve the generalization of LLMs' inference, especially when they contain dataset bias.
In the MRC experiments, the natural questions dataset used is constructed based on Wikipedia, which mainly consists of world facts.
This potentially allows for a simplicity bias of LLMs where questions can be answered without contexts.
Therefore, our study suggests the importance of using counterfactual demonstrations in knowledge conflict scenarios.
% Therefore, our study suggests the importance of considering the potential biases in the task data and carefully selecting demonstration instances to avoid dataset bias.

\subsection{Prediction with Abstention}\label{ssec:abstention}

\smallskip
\noindent\textbf{Datasets.} As for the second aspect of faithfulness, we evaluate LLMs' ability to selectively abstain from making uncertain predictions based on irrelevant context.
Since existing datasets such as SQuAD 2.0~\cite{rajpurkar-etal-2018-know} generally contain questions with confusion~\cite{yatskar-2019-qualitative} and are less related to our problem setting, we curate our own evaluation data based on RealTime QA~\cite{kasai2022realtime}, a dataset that inquires about novel information from June 2022 onwards.
In this formulation, LLMs are presented with a question and multiple choices, and they need to choose the correct answer based on several retrieved documents.
These documents were obtained using tools like Google custom search and may not contain the answer to the question.
To adapt this dataset to our setting, we added a new ``I don't know'' choice and relabeled the dataset. 
Instances where the retrieved documents do not answer the question are relabeled to ``I don't know''.
We used questions in the first six weeks of 2022 as the test set and randomly picked three questions of 2023 as demonstration instances.
This process results in a total of 113 test instances, including 63 answerable questions and 50 unanswerable ones.
% \muhao{may consider adding a footnote saying that SQuAD 2.0 is not about information after the training timestamp of GPT-3, so does not apply to the experiment even it provides UNK answers.}

\smallskip
\noindent\textbf{Task setup.}
We calculate the probability of a choice as $P(\text{choice}|\text{prompt})$ followed by normalization across all choices.\footnote{We tried three methods to calculate $P(\text{choice}|\text{prompt})$: joint probability, per-token probability (joint probability normalized by length), and unconditional probability as done in~\citet{brown2020language}. We find that joint probability works the best for GPT-3.5, while per-token probability works the best for LLama-2.}
% We take the choice with the largest probability as the prediction.
We report accuracy on the entire dataset (All), accuracy on the subset of questions that can be answered based on retrieved documents (HasAns), and accuracy on questions that cannot be answered based on retrieved documents (NoAns).
The latter two metrics measure LLMs' ability to extract answers from context and their ability to abstain from making predictions when the context does not describe the answer, respectively.
Besides, we use the probability of ``I don't know'' as LLM's probability estimation of whether the question can be answered.
We use the Brier score to evaluate the accuracy of the estimation, which measures the mean squared difference between the estimation and the true binary outcome of answerability.
We use three demonstrations for each instance in the few-shot setting, where some instances are filtered out during evaluation due to exceeding the context length of LLMs.

\smallskip
\noindent\textbf{Results and discussion.}
The results detailed in~\Cref{tab:realtime_qa_result} reveal that the \textsc{Opin + Instr} prompt outperforms all others on GPT-3.5, both in the zero-shot and few-shot settings, surpassing base prompts by 57.2\% and 16.3\% in NoAns subset accuracy, respectively.
For LLama-2, this approach similarly outperforms base prompts by 22.4\% in both settings.
Furthermore, the Brier score is reduced by 24.2\% and 7.8\% compared to base prompts for GPT-3.5 in the two settings, respectively, and by 2.6\% and 1.9\% on LLama-2.
The \textsc{Opin} prompt is the second best in terms of these metrics.
These findings demonstrate that opinion-based prompts can enhance the LLMs' ability to make selective predictions.
% Furthermore, all proposed prompting templates achieve better results on NoAns instances compared to base prompts, and maintain perfect accuracy on HasAns subset, indicating their ability to improve LLMs' selective prediction without compromising performance on answerable instances.
In addition, The use of demonstrations consistently improves the LLMs' ability to make selective predictions, as evidenced by the lower Brier scores and higher NoAns accuracy in the few-shot setting compared to the zero-shot setting.

% However, opinion-based prompts underperform base prompts on HasAns instances.
% We find that the large performance gap is mainly caused by wrong predictions of ``don't know''.
% When evaluated on HasAns instances where neither of the prompt types %predict none-``don't know'' answers
% lead to abstention, the underperformance is reduced to to 0.5\% in both zero-shot and few-shot settings.
% Therefore, these results suggest that opinion-based prompts are more suitable for scenarios requiring abstention.

\subsection{Additional Analysis}
\label{ssec:analysis}

\begin{figure}[!t]
    \centering
    \includegraphics[width=0.85\linewidth]{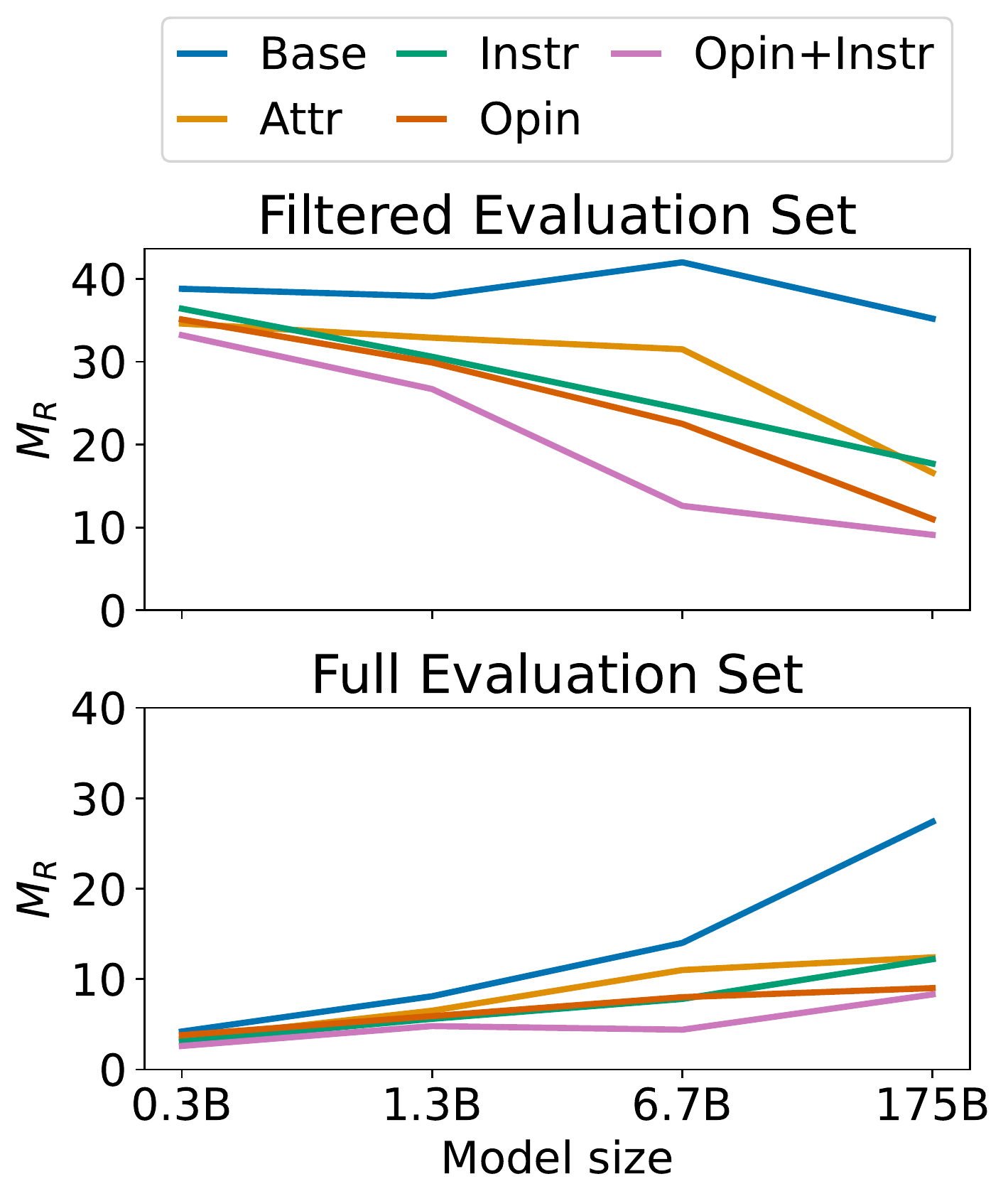}
    \caption{Memorization ratios across different sizes of InstructGPTs, evaluated in the zero-shot setting.}
    \label{fig:scaling}
    %\vspace{-1em}
\end{figure}

\smallskip
\noindent
\textbf{Memorization by different sizes of LLMs.} \Cref{fig:scaling} shows the memorization ratio $M_R$ across different sizes of InstructGPTs under the zero-shot evaluation of natural questions.\footnote{The 0.3B, 1.3B, 6.7B models refer to text-ada-001, text-babbage-001, text-curie-001, respectively. We do not perform few-shot evaluation as different sizes of LLMs have different maximum input lengths and can take different numbers of demonstrations, thus hard to be compared to each other.}
Overall, \textsc{Opin + Instr} consistently outperforms other prompts across different model sizes.
In the upper plot, results are shown for filtered evaluation sets where the corresponding LLMs can correctly predict the original answers without additional contexts, thereof the size of evaluation sets varies across different LLMs.\footnote{The sizes of the filtered evaluation sets, in the order of increased model sizes, are 121, 132, 756, and 2,773.}
We observe that $M_R$ generally decreases with increased model size, showing that larger LLMs are better at updating memorized answers based on given contexts in knowledge conflicts.
However, the lower plot reveals that larger LLMs have more severe memorization on the full (unfiltered) evaluation set.
This is because larger LLMs can memorize more answers than smaller ones, as evidenced by the number of instances in the filtered evaluation set where larger LLMs have more instances.
Our analysis suggests that while larger LLMs are better at updating memorized answers, they still tend to have more memorization due to the larger number of memorized answers.
Therefore, we need to pay more attention when using larger LLMs in %potentially knowledge conflict scenarios.
scenarios with new or potentially conflicting knowledge.

\begin{figure}
    \centering
    \includegraphics[width=0.85\linewidth]{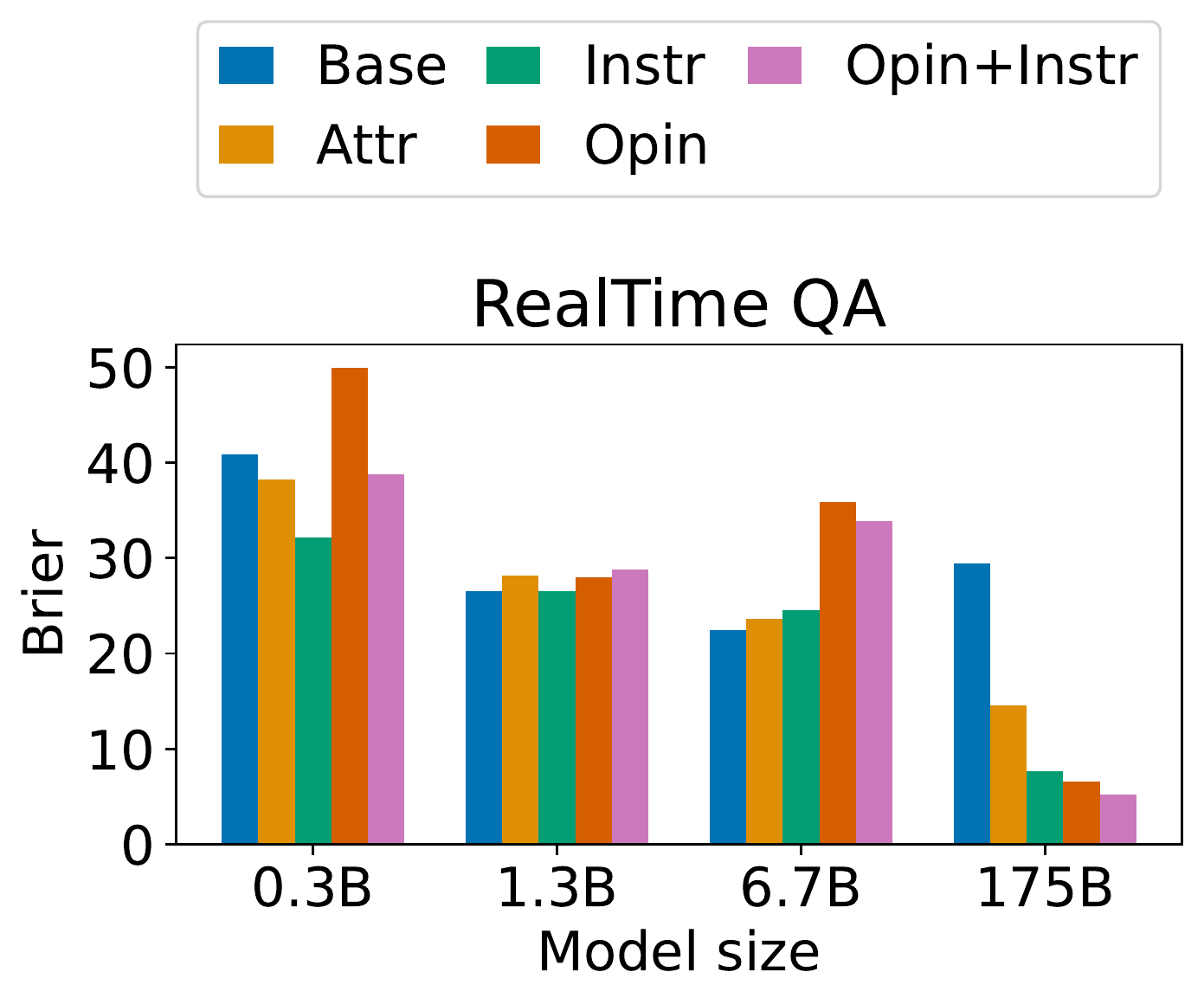}
    \caption{Brier scores across different sizes of InstructGPTs in the zero-shot setting of RealTime QA.}
    \label{fig:abstention}
    %\vspace{-1em}
\end{figure}

\smallskip
\noindent
\textbf{Selective prediction by different sizes of LLMs.}
\Cref{fig:abstention} shows the Brier score across different sizes of InstructGPTs under the zero-shot evaluation of RealTime QA.
On smaller LLMs, opinion-based prompt achieves similar or even higher Brier score than base prompts, indicating it does not improve the selective prediction ability of LLMs.
We hypothesize that this is because smaller LLMs have inferior reading comprehension ability, resulting in uncertainty in many instances.
Opinion-based prompts change uncertain predictions of answerable questions to \textit{I don't know}, which could lead to worse results.
For other prompting templates, we do not observe a consistent improvement across different LLMs either.
This analysis shows that while the selective prediction ability can be more easily activated by zero-shot prompting for LLMs such as text-davinci-003, smaller LLMs may require dedicated adaptations such as calibration and finetuning to activate this ability.

\smallskip
\noindent
\textbf{Results on original datasets.}
While our main experiments demonstrate the effectiveness of the proposed methods in resolving knowledge conflicts, LLMs in real-world applications may also see instances without knowledge conflicts.
Therefore, we investigate how our methods affect inference when the memorized answers align with the given contexts.
To do so, we evaluate LLMs on the same set of filtered evaluation set used in the main results section (\Cref{ssec:knowledge_conflict}), but we use the original contexts and answers instead of counterfactual ones.
The results in~\Cref{tab:original_data} show that opinion-based prompts yield similar or better results in all settings.
Furthermore, using either counterfactual or original demonstrations does not significantly impact results on the original (factual) dataset.
This analysis reveals that our methods do not impair performance on instances without knowledge conflicts.

\begin{table}[!t]
\setlength\extrarowheight{1pt}
    \centering
    %\scalebox{0.8}
    {\small
    \begin{tabular}{llcc}
    \toprule
     &Method& $p_o$$\uparrow$& EM$\uparrow$\\
     \midrule
     \multirow{3}{*}{\rotatebox[origin=c]{90}{Zero-shot}}&Base& 92.1& 11.1\\
     &Opin& 91.3& 25.2\\
     &Opin + Instr& 90.5& 57.2\\
     \midrule
     \multirow{3}{*}{\rotatebox[origin=c]{90}{Original}}&Base& 93.2& 77.8\\
     &Opin& 92.7& 78.7\\
     &Opin + Instr& 93.9& 80.1\\
     \midrule
     \multirow{3}{*}{\rotatebox[origin=c]{90}{Counter}}&Base& 93.6& 82.1\\
     &Opin& 92.8& 82.3\\
     &Opin + Instr& 92.7& 82.1\\
     \bottomrule
    \end{tabular}}
    \caption{Results (in \%) on the filtered evaluation set of natural questions with original contexts and answers.}
    \label{tab:original_data}
    %\vspace{-1em}
\end{table}

\begin{table*}[t!]
\setlength\extrarowheight{1pt}
    \centering
    \scalebox{0.81}
    {
    \begin{tabular}{lp{0.5\linewidth}p{0.5\linewidth}}
    \toprule
    & \textbf{Knowledge Conflict}& \textbf{Prediction with Abstention} \\
    \midrule
     \textbf{Context}& (Counterfactual passage) The Super Bowl LI Halftime show took place on February 5, 2017, at NRG Stadium in Houston, Texas as part of Super Bowl LI. The show was headlined by \textbf{Bosco}, who performed a medley of her songs, including newer material from her most recent studio album Joanne.& Tara Connolly is senior gas campaigner at Global Witness, an international NGO working towards a more sustainable, just and equal planet. She has over a decade of experience in EU energy policy. The views expressed in this commentary are her own.\\
     \midrule
     \textbf{Prompt}& \textcolor{orange}{Instruction: read the given information and answer the corresponding question.}& \textcolor{orange}{Instruction: answer a question based on the provided input-output pairs.}\\
     &\textcolor{cyan}{Bob said, ``}The Super Bowl ... album Joanne.\textcolor{cyan}{''}& \textcolor{cyan}{Bob said, ``}Tara Connolly ... are her own.\textcolor{cyan}{''} \\
     & Q: who performed the halftime show at Super Bowl 51 \textcolor{cyan}{in Bob's opinion} \textcolor{teal}{based on the given text}?& Q: Mo Farah made public that he was trafficked from which African country to the UK \textcolor{cyan}{in Bob's opinion} \textcolor{teal}{based on the given text}? \\
     &&Choices: Somaliland; Djibouti; Ethiopia; Somalia; I don't know\\
     \midrule
     \textbf{Base}& Lady Gaga \textcolor{red}{\ding{55}}& Somalia \textcolor{red}{\ding{55}}\\
     \textbf{Attr}& Lady Gaga \textcolor{red}{\ding{55}}& Somalia \textcolor{red}{\ding{55}} \\
     \textbf{Instr}&  Lady Gaga \textcolor{red}{\ding{55}}& Somaliland \textcolor{red}{\ding{55}}\\
     \textbf{Opin}&  Bosco \textcolor{GREEN}{\ding{51}}& I don't know \textcolor{GREEN}{\ding{51}}\\
     \textbf{Instr + Opin}& Bosco \textcolor{GREEN}{\ding{51}}& I don't know \textcolor{GREEN}{\ding{51}}\\
     \midrule
     \textbf{Answer}& Bosco&I don't know \\
     \bottomrule
    \end{tabular}}
    \caption{Examples of prompts and LLMs' corresponding predictions. In the ``Prompt'' row, we show and highlight the added parts from different prompting templates including \textcolor{teal}{attributed prompts}, \textcolor{orange}{instruction-based prompts}, and \textcolor{cyan}{opinion-based prompts}.}
    \label{tab:case_study}
    %\vspace{-1em}
\end{table*}

\subsection{Case Study}
\label{ssec:case_study}
\Cref{tab:case_study} shows some examples of prompts and the corresponding answers generated by text-davinci-003.
The left column of the table presents a knowledge conflict case where the original answer, \textit{Lady Gaga}, is replaced with a counterfactual answer, \textit{Bosco}.
When using base prompts, LLM ignores the context and return the memorized answer \textit{Lady Gaga}.
However, using opinion-based prompts and their combination with instructions leads to a more faithful response, with the language model returning \textit{Bosco} in the given context.
The right column presents a scenario where the retrieved context from Google search is irrelevant to the given question.
In such cases, base prompts still return a choice, leading to a potentially incorrect answer.
However, opinion-based prompts and their combination with instructions can abstain from making predictions and return \textit{I don't know}.
These examples demonstrate the effectiveness of proposed prompts in generating context-faithful responses.

\section{Conclusion}
In this paper, we focus on addressing the faithfulness issue of LLMs in context-specific NLP tasks, particularly in scenarios with knowledge conflict and prediction with abstention.
We propose that two methods, opinion-based prompts and counterfactual demonstrations, are effective in improving LLMs' faithfulness to contexts.
We evaluate our methods on three datasets of two tasks, namely machine reading comprehension and relation extraction, and observed significant improvement in faithfulness to contexts.
% Our additional analysis reveals that knowledge conflict is more severe for larger LLMs, and our proposed methods do not hurt performance on original datasets.
Future work includes evaluating the effectiveness of proposed methods on a broader range of NLP tasks such as open-domain QA and summarization, and studying other techniques to improve faithfulness further.

\section*{Acknowledgement}
We appreciate the reviewers for their insightful
comments and suggestions.
Wenxuan Zhou and Muhao Chen are supported by the NSF Grant IIS 2105329 and 
the DARPA MCS program under Contract No. N660011924033 with
the United States Office Of Naval Research.

\section*{Limitations}
In this study, our main focus is on the utilization of context-augmented prompting, assuming the reliability of the provided context. However, real-world scenarios can be more complicated, which may involve retrieved contexts that contain erroneous or conflicting information. Assessing the factuality of the context solely based on the provided information becomes challenging, as it depends on additional factors such as %source reliability or the timeliness of updates.
trustworthiness and timeliness of the information source.
Due to the complexity and challenges associated with verifying context reliability, we do not address this issue within the scope of this work. Furthermore, it is important to note that our paper primarily concentrates on the capability of LLMs to generate updated answers or decisions for given questions, rather than exploring more intricate tasks that require the model to apply the updated knowledge in multi-hop reasoning.

\section*{Ethical Considerations}
%The datasets used in this paper are exclusively in English, as they are only available in that language version.
Due to the availability of test data, the experiments conducted in this work has been in English, while future work can consider extending the use of proposed techniques to tasks in other languages.
%Besides, the utilization of public datasets introduces the possibility of inherent biases present in the data.
The datasets used in this work are public datasets that may not be free of inherent biases.
However, the introduced context-faithful prompting techniques in this work do not introduce additional biases beyond what the data have presented.

\bibliography{anthology,custom}

\begin{thebibliography}{56}
\expandafter\ifx\csname natexlab\endcsname\relax\def\natexlab#1{#1}\fi

\bibitem[{Bjerva et~al.(2020)Bjerva, Bhutani, Golshan, Tan, and
  Augenstein}]{bjerva-etal-2020-subjqa}
Johannes Bjerva, Nikita Bhutani, Behzad Golshan, Wang-Chiew Tan, and Isabelle
  Augenstein. 2020.
\newblock \href {https://doi.org/10.18653/v1/2020.emnlp-main.442} {{SubjQA}:
  {A} {D}ataset for {S}ubjectivity and {R}eview {C}omprehension}.
\newblock In \emph{Proceedings of the 2020 Conference on Empirical Methods in
  Natural Language Processing (EMNLP)}, pages 5480--5494, Online. Association
  for Computational Linguistics.

\bibitem[{Brown et~al.(2020)Brown, Mann, Ryder, Subbiah, Kaplan, Dhariwal,
  Neelakantan, Shyam, Sastry, Askell et~al.}]{brown2020language}
Tom Brown, Benjamin Mann, Nick Ryder, Melanie Subbiah, Jared~D Kaplan, Prafulla
  Dhariwal, Arvind Neelakantan, Pranav Shyam, Girish Sastry, Amanda Askell,
  et~al. 2020.
\newblock Language models are few-shot learners.
\newblock \emph{Advances in neural information processing systems},
  33:1877--1901.

\bibitem[{Choi et~al.(2018)Choi, He, Iyyer, Yatskar, Yih, Choi, Liang, and
  Zettlemoyer}]{choi-etal-2018-quac}
Eunsol Choi, He~He, Mohit Iyyer, Mark Yatskar, Wen-tau Yih, Yejin Choi, Percy
  Liang, and Luke Zettlemoyer. 2018.
\newblock \href {https://doi.org/10.18653/v1/D18-1241} {{Q}u{AC}: Question
  answering in context}.
\newblock In \emph{Proceedings of the 2018 Conference on Empirical Methods in
  Natural Language Processing}, pages 2174--2184, Brussels, Belgium.
  Association for Computational Linguistics.

\bibitem[{Chow(1970)}]{chow1970optimum}
C~Chow. 1970.
\newblock On optimum recognition error and reject tradeoff.
\newblock \emph{IEEE Transactions on information theory}, 16(1):41--46.

\bibitem[{Chowdhery et~al.(2022)Chowdhery, Narang, Devlin, Bosma, Mishra,
  Roberts, Barham, Chung, Sutton, Gehrmann et~al.}]{chowdhery2022palm}
Aakanksha Chowdhery, Sharan Narang, Jacob Devlin, Maarten Bosma, Gaurav Mishra,
  Adam Roberts, Paul Barham, Hyung~Won Chung, Charles Sutton, Sebastian
  Gehrmann, et~al. 2022.
\newblock Palm: Scaling language modeling with pathways.
\newblock \emph{arXiv preprint arXiv:2204.02311}.

\bibitem[{Chung et~al.(2022)Chung, Hou, Longpre, Zoph, Tay, Fedus, Li, Wang,
  Dehghani, Brahma et~al.}]{chung2022scaling}
Hyung~Won Chung, Le~Hou, Shayne Longpre, Barret Zoph, Yi~Tay, William Fedus,
  Eric Li, Xuezhi Wang, Mostafa Dehghani, Siddhartha Brahma, et~al. 2022.
\newblock Scaling instruction-finetuned language models.
\newblock \emph{arXiv preprint arXiv:2210.11416}.

\bibitem[{Clark et~al.(2019)Clark, Lee, Chang, Kwiatkowski, Collins, and
  Toutanova}]{clark-etal-2019-boolq}
Christopher Clark, Kenton Lee, Ming-Wei Chang, Tom Kwiatkowski, Michael
  Collins, and Kristina Toutanova. 2019.
\newblock \href {https://doi.org/10.18653/v1/N19-1300} {{B}ool{Q}: Exploring
  the surprising difficulty of natural yes/no questions}.
\newblock In \emph{Proceedings of the 2019 Conference of the North {A}merican
  Chapter of the Association for Computational Linguistics: Human Language
  Technologies, Volume 1 (Long and Short Papers)}, pages 2924--2936,
  Minneapolis, Minnesota. Association for Computational Linguistics.

\bibitem[{Clark et~al.(2018)Clark, Cowhey, Etzioni, Khot, Sabharwal, Schoenick,
  and Tafjord}]{clark2018think}
Peter Clark, Isaac Cowhey, Oren Etzioni, Tushar Khot, Ashish Sabharwal, Carissa
  Schoenick, and Oyvind Tafjord. 2018.
\newblock Think you have solved question answering? try arc, the ai2 reasoning
  challenge.
\newblock \emph{arXiv preprint arXiv:1803.05457}.

\bibitem[{Cortes et~al.(2016)Cortes, DeSalvo, and Mohri}]{cortes2016learning}
Corinna Cortes, Giulia DeSalvo, and Mehryar Mohri. 2016.
\newblock Learning with rejection.
\newblock In \emph{Algorithmic Learning Theory: 27th International Conference,
  ALT 2016, Bari, Italy, October 19-21, 2016, Proceedings 27}, pages 67--82.
  Springer.

\bibitem[{De~Cao et~al.(2021)De~Cao, Aziz, and
  Titov}]{de-cao-etal-2021-editing}
Nicola De~Cao, Wilker Aziz, and Ivan Titov. 2021.
\newblock \href {https://doi.org/10.18653/v1/2021.emnlp-main.522} {Editing
  factual knowledge in language models}.
\newblock In \emph{Proceedings of the 2021 Conference on Empirical Methods in
  Natural Language Processing}, pages 6491--6506, Online and Punta Cana,
  Dominican Republic. Association for Computational Linguistics.

\bibitem[{Fumera and Roli(2002)}]{fumera2002support}
Giorgio Fumera and Fabio Roli. 2002.
\newblock Support vector machines with embedded reject option.
\newblock In \emph{Pattern Recognition with Support Vector Machines: First
  International Workshop, SVM 2002 Niagara Falls, Canada, August 10, 2002
  Proceedings}, pages 68--82. Springer.

\bibitem[{Gal and Ghahramani(2016)}]{gal2016dropout}
Yarin Gal and Zoubin Ghahramani. 2016.
\newblock Dropout as a bayesian approximation: Representing model uncertainty
  in deep learning.
\newblock In \emph{international conference on machine learning}, pages
  1050--1059. PMLR.

\bibitem[{Gao and Callan(2022)}]{gao-callan-2022-unsupervised}
Luyu Gao and Jamie Callan. 2022.
\newblock \href {https://doi.org/10.18653/v1/2022.acl-long.203} {Unsupervised
  corpus aware language model pre-training for dense passage retrieval}.
\newblock In \emph{Proceedings of the 60th Annual Meeting of the Association
  for Computational Linguistics (Volume 1: Long Papers)}, pages 2843--2853,
  Dublin, Ireland. Association for Computational Linguistics.

\bibitem[{Gupta et~al.(2019)Gupta, Kulkarni, Chanda, Rayasam, and
  Lipton}]{guptaamazonqa19}
Mansi Gupta, Nitish Kulkarni, Raghuveer Chanda, Anirudha Rayasam, and Zachary~C
  Lipton. 2019.
\newblock Amazonqa: A review-based question answering task.
\newblock In \emph{International Joint Conference on Artificial Intelligence}.

\bibitem[{Hendrycks and Gimpel(2017)}]{hendrycksbaseline}
Dan Hendrycks and Kevin Gimpel. 2017.
\newblock A baseline for detecting misclassified and out-of-distribution
  examples in neural networks.
\newblock In \emph{International Conference on Learning Representations}.

\bibitem[{Hendrycks et~al.(2020)Hendrycks, Liu, Wallace, Dziedzic, Krishnan,
  and Song}]{hendrycks-etal-2020-pretrained}
Dan Hendrycks, Xiaoyuan Liu, Eric Wallace, Adam Dziedzic, Rishabh Krishnan, and
  Dawn Song. 2020.
\newblock \href {https://doi.org/10.18653/v1/2020.acl-main.244} {Pretrained
  transformers improve out-of-distribution robustness}.
\newblock In \emph{Proceedings of the 58th Annual Meeting of the Association
  for Computational Linguistics}, pages 2744--2751, Online. Association for
  Computational Linguistics.

\bibitem[{Izacard et~al.(2022)Izacard, Lewis, Lomeli, Hosseini, Petroni,
  Schick, Dwivedi-Yu, Joulin, Riedel, and Grave}]{izacard2022few}
Gautier Izacard, Patrick Lewis, Maria Lomeli, Lucas Hosseini, Fabio Petroni,
  Timo Schick, Jane Dwivedi-Yu, Armand Joulin, Sebastian Riedel, and Edouard
  Grave. 2022.
\newblock Few-shot learning with retrieval augmented language models.
\newblock \emph{arXiv preprint arXiv:2208.03299}.

\bibitem[{Joshi et~al.(2017)Joshi, Choi, Weld, and
  Zettlemoyer}]{joshi-etal-2017-triviaqa}
Mandar Joshi, Eunsol Choi, Daniel Weld, and Luke Zettlemoyer. 2017.
\newblock \href {https://doi.org/10.18653/v1/P17-1147} {{T}rivia{QA}: A large
  scale distantly supervised challenge dataset for reading comprehension}.
\newblock In \emph{Proceedings of the 55th Annual Meeting of the Association
  for Computational Linguistics (Volume 1: Long Papers)}, pages 1601--1611,
  Vancouver, Canada. Association for Computational Linguistics.

\bibitem[{Karpukhin et~al.(2020)Karpukhin, Oguz, Min, Lewis, Wu, Edunov, Chen,
  and Yih}]{karpukhin-etal-2020-dense}
Vladimir Karpukhin, Barlas Oguz, Sewon Min, Patrick Lewis, Ledell Wu, Sergey
  Edunov, Danqi Chen, and Wen-tau Yih. 2020.
\newblock \href {https://doi.org/10.18653/v1/2020.emnlp-main.550} {Dense
  passage retrieval for open-domain question answering}.
\newblock In \emph{Proceedings of the 2020 Conference on Empirical Methods in
  Natural Language Processing (EMNLP)}, pages 6769--6781, Online. Association
  for Computational Linguistics.

\bibitem[{Kasai et~al.(2022)Kasai, Sakaguchi, Takahashi, Bras, Asai, Yu, Radev,
  Smith, Choi, and Inui}]{kasai2022realtime}
Jungo Kasai, Keisuke Sakaguchi, Yoichi Takahashi, Ronan~Le Bras, Akari Asai,
  Xinyan Yu, Dragomir Radev, Noah~A Smith, Yejin Choi, and Kentaro Inui. 2022.
\newblock Realtime qa: What's the answer right now?
\newblock \emph{arXiv preprint arXiv:2207.13332}.

\bibitem[{Khattab et~al.(2022)Khattab, Santhanam, Li, Hall, Liang, Potts, and
  Zaharia}]{khattab2022demonstrate}
Omar Khattab, Keshav Santhanam, Xiang~Lisa Li, David Hall, Percy Liang,
  Christopher Potts, and Matei Zaharia. 2022.
\newblock Demonstrate-search-predict: Composing retrieval and language models
  for knowledge-intensive nlp.
\newblock \emph{arXiv preprint arXiv:2212.14024}.

\bibitem[{Kwiatkowski et~al.(2019)Kwiatkowski, Palomaki, Redfield, Collins,
  Parikh, Alberti, Epstein, Polosukhin, Devlin, Lee, Toutanova, Jones, Kelcey,
  Chang, Dai, Uszkoreit, Le, and Petrov}]{kwiatkowski-etal-2019-natural}
Tom Kwiatkowski, Jennimaria Palomaki, Olivia Redfield, Michael Collins, Ankur
  Parikh, Chris Alberti, Danielle Epstein, Illia Polosukhin, Jacob Devlin,
  Kenton Lee, Kristina Toutanova, Llion Jones, Matthew Kelcey, Ming-Wei Chang,
  Andrew~M. Dai, Jakob Uszkoreit, Quoc Le, and Slav Petrov. 2019.
\newblock \href {https://doi.org/10.1162/tacl_a_00276} {Natural questions: A
  benchmark for question answering research}.
\newblock \emph{Transactions of the Association for Computational Linguistics},
  7:452--466.

\bibitem[{Lakshminarayanan et~al.(2017)Lakshminarayanan, Pritzel, and
  Blundell}]{lakshminarayanan2017simple}
Balaji Lakshminarayanan, Alexander Pritzel, and Charles Blundell. 2017.
\newblock Simple and scalable predictive uncertainty estimation using deep
  ensembles.
\newblock \emph{Advances in neural information processing systems}, 30.

\bibitem[{Lazaridou et~al.(2022)Lazaridou, Gribovskaya, Stokowiec, and
  Grigorev}]{lazaridou2022internet}
Angeliki Lazaridou, Elena Gribovskaya, Wojciech Stokowiec, and Nikolai
  Grigorev. 2022.
\newblock Internet-augmented language models through few-shot prompting for
  open-domain question answering.
\newblock \emph{arXiv preprint arXiv:2203.05115}.

\bibitem[{Lazaridou et~al.(2021)Lazaridou, Kuncoro, Gribovskaya, Agrawal,
  Liska, Terzi, Gimenez, de~Masson~d'Autume, Kocisky, Ruder
  et~al.}]{lazaridou2021mind}
Angeliki Lazaridou, Adhi Kuncoro, Elena Gribovskaya, Devang Agrawal, Adam
  Liska, Tayfun Terzi, Mai Gimenez, Cyprien de~Masson~d'Autume, Tomas Kocisky,
  Sebastian Ruder, et~al. 2021.
\newblock Mind the gap: Assessing temporal generalization in neural language
  models.
\newblock \emph{Advances in Neural Information Processing Systems},
  34:29348--29363.

\bibitem[{Li et~al.(2022)Li, Rawat, Zaheer, Wang, Lukasik, Veit, Yu, and
  Kumar}]{li2022large}
Daliang Li, Ankit~Singh Rawat, Manzil Zaheer, Xin Wang, Michal Lukasik, Andreas
  Veit, Felix Yu, and Sanjiv Kumar. 2022.
\newblock Large language models with controllable working memory.
\newblock \emph{arXiv preprint arXiv:2211.05110}.

\bibitem[{Lin et~al.(2022)Lin, Hilton, and Evans}]{lin-etal-2022-truthfulqa}
Stephanie Lin, Jacob Hilton, and Owain Evans. 2022.
\newblock \href {https://doi.org/10.18653/v1/2022.acl-long.229}
  {{T}ruthful{QA}: Measuring how models mimic human falsehoods}.
\newblock In \emph{Proceedings of the 60th Annual Meeting of the Association
  for Computational Linguistics (Volume 1: Long Papers)}, pages 3214--3252,
  Dublin, Ireland. Association for Computational Linguistics.

\bibitem[{Liska et~al.(2022)Liska, Kocisky, Gribovskaya, Terzi, Sezener,
  Agrawal, Cyprien De~Masson, Scholtes, Zaheer, Young
  et~al.}]{liska2022streamingqa}
Adam Liska, Tomas Kocisky, Elena Gribovskaya, Tayfun Terzi, Eren Sezener,
  Devang Agrawal, D’Autume Cyprien De~Masson, Tim Scholtes, Manzil Zaheer,
  Susannah Young, et~al. 2022.
\newblock Streamingqa: A benchmark for adaptation to new knowledge over time in
  question answering models.
\newblock In \emph{International Conference on Machine Learning}, pages
  13604--13622. PMLR.

\bibitem[{Liu et~al.(2022)Liu, Shen, Zhang, Dolan, Carin, and
  Chen}]{liu-etal-2022-makes}
Jiachang Liu, Dinghan Shen, Yizhe Zhang, Bill Dolan, Lawrence Carin, and Weizhu
  Chen. 2022.
\newblock \href {https://doi.org/10.18653/v1/2022.deelio-1.10} {What makes good
  in-context examples for {GPT}-3?}
\newblock In \emph{Proceedings of Deep Learning Inside Out (DeeLIO 2022): The
  3rd Workshop on Knowledge Extraction and Integration for Deep Learning
  Architectures}, pages 100--114, Dublin, Ireland and Online. Association for
  Computational Linguistics.

\bibitem[{Liu et~al.(2019)Liu, Ott, Goyal, Du, Joshi, Chen, Levy, Lewis,
  Zettlemoyer, and Stoyanov}]{liu2019roberta}
Yinhan Liu, Myle Ott, Naman Goyal, Jingfei Du, Mandar Joshi, Danqi Chen, Omer
  Levy, Mike Lewis, Luke Zettlemoyer, and Veselin Stoyanov. 2019.
\newblock Roberta: A robustly optimized bert pretraining approach.
\newblock \emph{arXiv preprint arXiv:1907.11692}.

\bibitem[{Longpre et~al.(2021)Longpre, Perisetla, Chen, Ramesh, DuBois, and
  Singh}]{longpre-etal-2021-entity}
Shayne Longpre, Kartik Perisetla, Anthony Chen, Nikhil Ramesh, Chris DuBois,
  and Sameer Singh. 2021.
\newblock \href {https://doi.org/10.18653/v1/2021.emnlp-main.565} {Entity-based
  knowledge conflicts in question answering}.
\newblock In \emph{Proceedings of the 2021 Conference on Empirical Methods in
  Natural Language Processing}, pages 7052--7063, Online and Punta Cana,
  Dominican Republic. Association for Computational Linguistics.

\bibitem[{Lu et~al.(2022)Lu, Hsu, Zhou, Ma, and
  Chen}]{lu-etal-2022-summarization}
Keming Lu, I-Hung Hsu, Wenxuan Zhou, Mingyu~Derek Ma, and Muhao Chen. 2022.
\newblock \href {https://doi.org/10.18653/v1/2022.findings-emnlp.490}
  {Summarization as indirect supervision for relation extraction}.
\newblock In \emph{Findings of the Association for Computational Linguistics:
  EMNLP 2022}, pages 6575--6594, Abu Dhabi, United Arab Emirates. Association
  for Computational Linguistics.

\bibitem[{Meng et~al.(2022)Meng, Bau, Andonian, and
  Belinkov}]{meng2022locating}
Kevin Meng, David Bau, Alex~J Andonian, and Yonatan Belinkov. 2022.
\newblock Locating and editing factual associations in gpt.
\newblock In \emph{Advances in Neural Information Processing Systems}.

\bibitem[{Meng et~al.(2023)Meng, Sharma, Andonian, Belinkov, and
  Bau}]{meng2022mass}
Kevin Meng, Arnab~Sen Sharma, Alex Andonian, Yonatan Belinkov, and David Bau.
  2023.
\newblock Mass-editing memory in a transformer.
\newblock In \emph{International Conference on Learning Representations}.

\bibitem[{Mihaylov et~al.(2018)Mihaylov, Clark, Khot, and
  Sabharwal}]{mihaylov-etal-2018-suit}
Todor Mihaylov, Peter Clark, Tushar Khot, and Ashish Sabharwal. 2018.
\newblock \href {https://doi.org/10.18653/v1/D18-1260} {Can a suit of armor
  conduct electricity? a new dataset for open book question answering}.
\newblock In \emph{Proceedings of the 2018 Conference on Empirical Methods in
  Natural Language Processing}, pages 2381--2391, Brussels, Belgium.
  Association for Computational Linguistics.

\bibitem[{Mitchell et~al.(2022)Mitchell, Lin, Bosselut, Finn, and
  Manning}]{mitchellfast}
Eric Mitchell, Charles Lin, Antoine Bosselut, Chelsea Finn, and Christopher~D
  Manning. 2022.
\newblock Fast model editing at scale.
\newblock In \emph{International Conference on Learning Representations}.

\bibitem[{Neeman et~al.(2022)Neeman, Aharoni, Honovich, Choshen, Szpektor, and
  Abend}]{neeman2022disentqa}
Ella Neeman, Roee Aharoni, Or~Honovich, Leshem Choshen, Idan Szpektor, and Omri
  Abend. 2022.
\newblock Disentqa: Disentangling parametric and contextual knowledge with
  counterfactual question answering.
\newblock \emph{arXiv preprint arXiv:2211.05655}.

\bibitem[{Rajpurkar et~al.(2018)Rajpurkar, Jia, and
  Liang}]{rajpurkar-etal-2018-know}
Pranav Rajpurkar, Robin Jia, and Percy Liang. 2018.
\newblock \href {https://doi.org/10.18653/v1/P18-2124} {Know what you don{'}t
  know: Unanswerable questions for {SQ}u{AD}}.
\newblock In \emph{Proceedings of the 56th Annual Meeting of the Association
  for Computational Linguistics (Volume 2: Short Papers)}, pages 784--789,
  Melbourne, Australia. Association for Computational Linguistics.

\bibitem[{Rajpurkar et~al.(2016)Rajpurkar, Zhang, Lopyrev, and
  Liang}]{rajpurkar-etal-2016-squad}
Pranav Rajpurkar, Jian Zhang, Konstantin Lopyrev, and Percy Liang. 2016.
\newblock \href {https://doi.org/10.18653/v1/D16-1264} {{SQ}u{AD}: 100,000+
  questions for machine comprehension of text}.
\newblock In \emph{Proceedings of the 2016 Conference on Empirical Methods in
  Natural Language Processing}, pages 2383--2392, Austin, Texas. Association
  for Computational Linguistics.

\bibitem[{Reddy et~al.(2019)Reddy, Chen, and Manning}]{reddy-etal-2019-coqa}
Siva Reddy, Danqi Chen, and Christopher~D. Manning. 2019.
\newblock \href {https://doi.org/10.1162/tacl_a_00266} {{C}o{QA}: A
  conversational question answering challenge}.
\newblock \emph{Transactions of the Association for Computational Linguistics},
  7:249--266.

\bibitem[{Reimers and Gurevych(2019)}]{reimers-gurevych-2019-sentence}
Nils Reimers and Iryna Gurevych. 2019.
\newblock \href {https://doi.org/10.18653/v1/D19-1410} {Sentence-{BERT}:
  Sentence embeddings using {S}iamese {BERT}-networks}.
\newblock In \emph{Proceedings of the 2019 Conference on Empirical Methods in
  Natural Language Processing and the 9th International Joint Conference on
  Natural Language Processing (EMNLP-IJCNLP)}, pages 3982--3992, Hong Kong,
  China. Association for Computational Linguistics.

\bibitem[{Sang and De~Meulder(2003)}]{sang2003introduction}
Erik~F Sang and Fien De~Meulder. 2003.
\newblock Introduction to the conll-2003 shared task: Language-independent
  named entity recognition.
\newblock \emph{arXiv preprint cs/0306050}.

\bibitem[{Santhanam et~al.(2022)Santhanam, Khattab, Saad-Falcon, Potts, and
  Zaharia}]{santhanam-etal-2022-colbertv2}
Keshav Santhanam, Omar Khattab, Jon Saad-Falcon, Christopher Potts, and Matei
  Zaharia. 2022.
\newblock \href {https://doi.org/10.18653/v1/2022.naacl-main.272}
  {{C}ol{BERT}v2: Effective and efficient retrieval via lightweight late
  interaction}.
\newblock In \emph{Proceedings of the 2022 Conference of the North American
  Chapter of the Association for Computational Linguistics: Human Language
  Technologies}, pages 3715--3734, Seattle, United States. Association for
  Computational Linguistics.

\bibitem[{Si et~al.(2023)Si, Gan, Yang, Wang, Wang, Boyd-Graber, and
  Wang}]{si2022prompting}
Chenglei Si, Zhe Gan, Zhengyuan Yang, Shuohang Wang, Jianfeng Wang, Jordan
  Boyd-Graber, and Lijuan Wang. 2023.
\newblock Prompting gpt-3 to be reliable.
\newblock In \emph{International Conference on Learning Representations}.

\bibitem[{Stoica et~al.(2021)Stoica, Platanios, and P{\'o}czos}]{stoica2021re}
George Stoica, Emmanouil~Antonios Platanios, and Barnab{\'a}s P{\'o}czos. 2021.
\newblock Re-tacred: Addressing shortcomings of the tacred dataset.
\newblock In \emph{Proceedings of the AAAI Conference on Artificial
  Intelligence}, volume~35, pages 13843--13850.

\bibitem[{Touvron et~al.(2023)Touvron, Martin, Stone, Albert, Almahairi,
  Babaei, Bashlykov, Batra, Bhargava, Bhosale et~al.}]{touvron2023llama}
Hugo Touvron, Louis Martin, Kevin Stone, Peter Albert, Amjad Almahairi, Yasmine
  Babaei, Nikolay Bashlykov, Soumya Batra, Prajjwal Bhargava, Shruti Bhosale,
  et~al. 2023.
\newblock Llama 2: Open foundation and fine-tuned chat models.
\newblock \emph{arXiv preprint arXiv:2307.09288}.

\bibitem[{Wang et~al.(2023)Wang, Zhang, Deng, Gardner, Roth, and
  Chen}]{wang-etal-2023-extracting}
Haoyu Wang, Hongming Zhang, Yuqian Deng, Jacob Gardner, Dan Roth, and Muhao
  Chen. 2023.
\newblock \href {https://doi.org/10.18653/v1/2023.eacl-main.39} {Extracting or
  guessing? improving faithfulness of event temporal relation extraction}.
\newblock In \emph{Proceedings of the 17th Conference of the European Chapter
  of the Association for Computational Linguistics}, pages 541--553, Dubrovnik,
  Croatia. Association for Computational Linguistics.

\bibitem[{Wang et~al.(2022)Wang, Chen, Zhou, Cai, Liang, Liu, Yang, Liu, and
  Hooi}]{wang-etal-2022-rely}
Yiwei Wang, Muhao Chen, Wenxuan Zhou, Yujun Cai, Yuxuan Liang, Dayiheng Liu,
  Baosong Yang, Juncheng Liu, and Bryan Hooi. 2022.
\newblock \href {https://doi.org/10.18653/v1/2022.naacl-main.224} {Should we
  rely on entity mentions for relation extraction? debiasing relation
  extraction with counterfactual analysis}.
\newblock In \emph{Proceedings of the 2022 Conference of the North American
  Chapter of the Association for Computational Linguistics: Human Language
  Technologies}, pages 3071--3081, Seattle, United States. Association for
  Computational Linguistics.

\bibitem[{Wei et~al.(2022)Wei, Bosma, Zhao, Guu, Yu, Lester, Du, Dai, and
  Le}]{weifinetuned}
Jason Wei, Maarten Bosma, Vincent Zhao, Kelvin Guu, Adams~Wei Yu, Brian Lester,
  Nan Du, Andrew~M Dai, and Quoc~V Le. 2022.
\newblock Finetuned language models are zero-shot learners.
\newblock In \emph{International Conference on Learning Representations}.

\bibitem[{Xin et~al.(2021)Xin, Tang, Yu, and Lin}]{xin-etal-2021-art}
Ji~Xin, Raphael Tang, Yaoliang Yu, and Jimmy Lin. 2021.
\newblock \href {https://doi.org/10.18653/v1/2021.acl-long.84} {The art of
  abstention: Selective prediction and error regularization for natural
  language processing}.
\newblock In \emph{Proceedings of the 59th Annual Meeting of the Association
  for Computational Linguistics and the 11th International Joint Conference on
  Natural Language Processing (Volume 1: Long Papers)}, pages 1040--1051,
  Online. Association for Computational Linguistics.

\bibitem[{Yatskar(2019)}]{yatskar-2019-qualitative}
Mark Yatskar. 2019.
\newblock \href {https://doi.org/10.18653/v1/N19-1241} {A qualitative
  comparison of {C}o{QA}, {SQ}u{AD} 2.0 and {Q}u{AC}}.
\newblock In \emph{Proceedings of the 2019 Conference of the North {A}merican
  Chapter of the Association for Computational Linguistics: Human Language
  Technologies, Volume 1 (Long and Short Papers)}, pages 2318--2323,
  Minneapolis, Minnesota. Association for Computational Linguistics.

\bibitem[{Zhang et~al.(2017)Zhang, Zhong, Chen, Angeli, and
  Manning}]{zhang-etal-2017-position}
Yuhao Zhang, Victor Zhong, Danqi Chen, Gabor Angeli, and Christopher~D.
  Manning. 2017.
\newblock \href {https://doi.org/10.18653/v1/D17-1004} {Position-aware
  attention and supervised data improve slot filling}.
\newblock In \emph{Proceedings of the 2017 Conference on Empirical Methods in
  Natural Language Processing}, pages 35--45, Copenhagen, Denmark. Association
  for Computational Linguistics.

\bibitem[{Zhou and Chen(2022)}]{zhou-chen-2022-improved}
Wenxuan Zhou and Muhao Chen. 2022.
\newblock \href {https://aclanthology.org/2022.aacl-short.21} {An improved
  baseline for sentence-level relation extraction}.
\newblock In \emph{Proceedings of the 2nd Conference of the Asia-Pacific
  Chapter of the Association for Computational Linguistics and the 12th
  International Joint Conference on Natural Language Processing (Volume 2:
  Short Papers)}, pages 161--168, Online only. Association for Computational
  Linguistics.

\bibitem[{Zhou et~al.(2021)Zhou, Liu, and Chen}]{zhou-etal-2021-contrastive}
Wenxuan Zhou, Fangyu Liu, and Muhao Chen. 2021.
\newblock \href {https://doi.org/10.18653/v1/2021.emnlp-main.84} {Contrastive
  out-of-distribution detection for pretrained transformers}.
\newblock In \emph{Proceedings of the 2021 Conference on Empirical Methods in
  Natural Language Processing}, pages 1100--1111, Online and Punta Cana,
  Dominican Republic. Association for Computational Linguistics.

\bibitem[{Zhou et~al.(2022)Zhou, Muresanu, Han, Paster, Pitis, Chan, and
  Ba}]{zhou2022large}
Yongchao Zhou, Andrei~Ioan Muresanu, Ziwen Han, Keiran Paster, Silviu Pitis,
  Harris Chan, and Jimmy Ba. 2022.
\newblock Large language models are human-level prompt engineers.
\newblock \emph{arXiv preprint arXiv:2211.01910}.

\bibitem[{Zhu et~al.(2020)Zhu, Rawat, Zaheer, Bhojanapalli, Li, Yu, and
  Kumar}]{zhu2020modifying}
Chen Zhu, Ankit~Singh Rawat, Manzil Zaheer, Srinadh Bhojanapalli, Daliang Li,
  Felix Yu, and Sanjiv Kumar. 2020.
\newblock Modifying memories in transformer models.
\newblock \emph{arXiv preprint arXiv:2012.00363}.

\end{thebibliography}
\bibliographystyle{acl_natbib}

\appendix
\begin{center}
    {
    \Large\textbf{Appendices}
    }
\end{center}
\section{Settings of Automatic Prompt Engineering}
\label{ssec:ape_instructions}
We run APE using their official code\footnote{\url{https://github.com/keirp/automatic_prompt_engineer}} and default hyperparameters.
In the knowledge conflict setting, we use counterfactual datasets to generate instructions.
While the APE paper recommends using instructions generated by the same model in inference, we find that smaller LLMs do not generate meaningful instructions for our datasets.
Therefore, we use instructions generated by text-davinci-003 across different scales of LLMs in additional analysis.
The top three instructions generated by APE on each dataset are listed below.
We use the top one instruction in experiments.

\smallskip
\noindent \textbf{Natural questions:}
\begin{enumerate}
    \item read the given information and answer the corresponding question.
    \item read a piece of text and then use the information in the text to answer a question.
    \item "Read the given information and answer the questions that follow."
\end{enumerate}
\smallskip
\noindent \textbf{Re-TACRED:}
\begin{enumerate}
    \item identify the relationship between two entities from a list of options.
    \item identify the relationship between two entities based on the given input-output pairs.
    \item identify the relationship between two entities given the input-output pairs.
\end{enumerate}
\smallskip
\noindent \textbf{RealTime QA:}
\begin{enumerate}
    \item answer a question based on the provided input-output pairs.
    \item ask a question with a set of choices and ask the friend to provide the correct answer.
    \item answer a question related to a news article.
\end{enumerate}
\end{document}